\documentclass[10pt,twocolumn,letterpaper]{article}

\usepackage{cvpr}
\usepackage{times}
\usepackage{epsfig}
\usepackage{graphicx}
\usepackage{amsmath}
\usepackage{amssymb}
\usepackage{multirow}
\usepackage{booktabs}
\usepackage{lettrine}
\usepackage{color}
\usepackage{subcaption}
\usepackage{url}

\usepackage[pagebackref=true,breaklinks=true,letterpaper=true,colorlinks,bookmarks=false]{hyperref}

\newcommand{\keypoint}[1]{\vspace{0.1cm}\noindent\textbf{#1}\quad}

\cvprfinalcopy 

\def\cvprPaperID{431} 

\ifcvprfinal\fi
\begin{document}

\title{Learning to Compare: Relation Network for Few-Shot Learning}

\author{Flood Sung \quad Yongxin Yang$^3$  \quad Li Zhang$^2$ \quad  Tao Xiang$^{1}$ \quad   Philip H.S. Torr$^{2}$  \quad  Timothy M. Hospedales$^{3}$\\
$^1$Queen Mary University of London \quad $^2$University of Oxford \quad $^3$The University of Edinburgh\\
{\tt\small floodsung@gmail.com \quad t.xiang@qmul.ac.uk} \\ 
{\tt\small \{lz, phst\}@robots.ox.ac.uk \quad \{yongxin.yang, t.hospedales\}@ed.ac.uk}
}

\maketitle
\thispagestyle{empty}

\begin{abstract}

We present a conceptually simple, flexible, and general framework for few-shot learning, where a classifier must learn to recognise new classes given only few examples from each. Our method, called the Relation Network (RN), is trained end-to-end from scratch. During meta-learning, it learns to learn a deep distance metric to compare a small number of images within episodes, each of which is designed to simulate the few-shot setting. 
Once trained, a RN is able to classify images of new classes by computing relation scores between query images and the few examples of each new class without further updating the network. 
Besides providing improved performance on few-shot learning, our framework is easily extended to zero-shot learning. Extensive experiments on five benchmarks demonstrate that our simple approach provides a unified and effective approach for both of these two tasks. 
\end{abstract}

\section{Introduction}
Deep learning models have achieved great success in visual recognition tasks \cite{krizhevsky2012imagenet, he2016deep, simonyan2014very}. However, these supervised learning models need large amounts of labelled data and many iterations to train their large number of parameters. This severely limits their scalability to new classes due to annotation cost, but more fundamentally limits their applicability to newly emerging (eg. new consumer devices) or rare (eg. rare animals) categories where numerous annotated images may simply never exist. In contrast, humans are very good at recognising objects with very little direct supervision, or none at all $\ie$, few-shot  \cite{lake2011one, fei2006one} or zero-shot  \cite{lampert2014attribute} learning. For example, children have no problem generalising the concept of ``zebra" from a single picture in a book, or hearing its description as looking like a stripy horse. Motivated by the failure of conventional deep learning methods to work well on one or few examples per class, and inspired by the few- and zero-shot learning ability of humans, there has been a recent resurgence of interest in machine one/few-shot \cite{edwards2016towards, vinyals2016matching, santoro2016meta, kaiser2017learning, koch2015siamese, finn2017model, munkhdalai2017meta, snell2017prototypical, ravi2016optimization} and zero-shot \cite{frome2013devise,akata2015evaluation,lampert2014attribute,zhang2017learning,lei2015predicting,romera2015embarrassingly} learning.

Few-shot learning aims to recognise novel visual categories from very few labelled examples. The availability of only one or very few examples challenges the standard `fine-tuning' practice in deep learning \cite{finn2017model}. Data augmentation and regularisation techniques can alleviate overfitting in such a limited-data regime, but they do not solve it. Therefore contemporary approaches to few-shot learning often decompose training into an auxiliary meta learning phase where transferrable knowledge is learned in the form of good initial conditions \cite{finn2017model}, embeddings \cite{snell2017prototypical,vinyals2016matching} or optimisation strategies \cite{ravi2016optimization}. The target few-shot learning problem is then learned by fine-tuning \cite{finn2017model} with the learned optimisation strategy \cite{ravi2016optimization} or computed in a feed-forward pass \cite{snell2017prototypical,vinyals2016matching,bertinetto2016feedForwardOneShot,santoro2016meta} without updating network weights. Zero-shot learning also suffers from a related challenge. Recognisers are trained by a single example in the form of a class description (c.f., single exemplar image in one-shot), making data insufficiency for gradient-based learning a challenge.

While promising, most existing few-shot learning approaches either require complex inference mechanisms \cite{lake2011one, fei2006one}, complex recurrent neural network (RNN) architectures \cite{vinyals2016matching,santoro2016meta}, or fine-tuning the target problem \cite{finn2017model,ravi2016optimization}. Our approach is most related to others that aim to train an effective metric for one-shot learning \cite{vinyals2016matching,snell2017prototypical,koch2015siamese}. Where they focus on the learning of the transferrable embedding and {pre-define a fixed metric} (e.g., as Euclidean \cite{snell2017prototypical}), we further aim to \emph{learn} a transferrable deep metric for comparing the relation between images (few-shot learning), or between images and class descriptions (zero-shot learning). 
By expressing the inductive bias of a \emph{deeper} solution (multiple non-linear learned stages at both embedding and relation modules), we make it easier to learn a generalisable solution to the problem. 


Specifically, we propose a two-branch Relation Network (RN) that performs few-shot recognition by learning to compare \emph{query} images against few-shot labeled {\em sample} images. First an {\em embedding module} generates representations of the query and training images. Then these embeddings are compared by a {\em relation module} that determines if they are from matching categories or not. Defining an episode-based strategy inspired by \cite{vinyals2016matching,snell2017prototypical}, the embedding and relation modules are meta-learned end-to-end to support few-shot learning. This can be seen as extending the strategy of \cite{vinyals2016matching,snell2017prototypical} to include a \emph{learnable non-linear} comparator, instead of a fixed linear comparator.  Our approach outperforms prior approaches, while being simpler (no RNNs \cite{vinyals2016matching,santoro2016meta,ravi2016optimization}) and faster  (no fine-tuning  \cite{ravi2016optimization,finn2017model}).
Our proposed strategy also directly generalises to zero-shot learning. In this case the sample branch embeds a single-shot category description rather than a single exemplar training image, and the relation module learns to compare query image and category description embeddings.

Overall our contribution is to provide a clean framework that elegantly encompasses both few and zero-shot learning. Our evaluation on four benchmarks show that it provides compelling performance across the board while being simpler and faster than the alternatives.

\section{Related Work}

The study of one or few-shot object recognition has been of interest for some time \cite{fei2006one}. Earlier work on few-shot learning tended to involve generative models with complex iterative inference strategies \cite{fei2006one,lake2011one}. With the success of discriminative deep learning-based approaches in the data-rich many-shot setting \cite{krizhevsky2012imagenet, he2016deep, simonyan2014very}, there has been a surge of interest in generalising  such deep learning approaches to the few-shot learning setting. Many of these approaches use a meta-learning or learning-to-learn strategy in the sense that they extract some transferrable knowledge from a set of auxiliary tasks  (meta-learning, learning-to-learn), which then helps them to learn the target few-shot problem well without suffering from the overfitting that might be expected when applying deep models to sparse data problems. 

\keypoint{Learning to Fine-Tune} The successful MAML approach \cite{finn2017model} aimed to meta-learn an initial condition (set of neural network weights) that is good for fine-tuning on few-shot problems. The strategy here is to search for the weight configuration of a given neural network such that it can be effectively fine-tuned on a sparse data problem within a few gradient-descent update steps. Many distinct target problems are sampled from a multiple task training set; the base neural network model is then fine-tuned to solve each of them, and the success at each target problem after fine-tuning drives updates in the base model -- thus driving the production of an easy to fine-tune initial condition. The few-shot optimisation approach \cite{ravi2016optimization} goes further in meta-learning not only a good initial condition but an LSTM-based optimizer that is trained to be specifically effective for fine-tuning. However both of these approaches suffer from the need to fine-tune on the target problem. In contrast, our approach solves target problems in an entirely feed-forward manner with no model updates required, making it more convenient for low-latency or low-power applications. 


\keypoint{RNN Memory Based} Another category of approaches leverage recurrent neural networks with memories \cite{munkhdalai2017meta,santoro2016meta}. Here the idea is typically that an RNN iterates over an examples of given problem and accumulates the knowledge required to solve that problem in its hidden activations, or external memory. New examples can be classified, for example by comparing them to historic information stored in the memory. So `learning' a  single target problem can occur in unrolling the RNN, while learning-to-learn means training the weights of the RNN by learning many distinct problems. While appealing, these architectures face issues in ensuring that they reliably store all the, potentially long term, historical information of relevance without forgetting. In our approach we avoid the complexity of recurrent networks, and the issues involved in ensuring the adequacy of their memory. Instead our learning-to-learn approach is defined entirely with simple and fast feed forward CNNs. 


\keypoint{Embedding and Metric Learning Approaches} The prior approaches entail some complexity when learning the target few-shot problem. Another category of approach aims to learn a set of projection functions that take query and sample images from the target problem and classify them in a feed forward manner \cite{vinyals2016matching,snell2017prototypical,bertinetto2016feedForwardOneShot}.  One approach is to parameterise the weights of a feed-forward classifier in terms of the sample set \cite{bertinetto2016feedForwardOneShot}. The meta-learning here is to train the auxiliary parameterisation net that learns how to paramaterise a given feed-forward classification problem in terms of a few-shot sample set. Metric-learning based approaches aim to learn a set of projection functions such that when represented in this embedding, images are easy to recognise using simple nearest neighbour or linear classifiers \cite{vinyals2016matching,snell2017prototypical,koch2015siamese}. In this case the meta-learned transferrable knowledge are the projection functions and the target problem is a simple feed-forward computation.

The most related methodologies to ours are the prototypical networks of \cite{snell2017prototypical} and the siamese networks of \cite{koch2015siamese}. These approaches focus on learning embeddings that transform the data such that it can be recognised with a \emph{fixed} nearest-neighbour \cite{snell2017prototypical} or linear \cite{koch2015siamese,snell2017prototypical} classifier. In contrast, our framework further defines a relation classifier CNN, in the style of \cite{santoro2017simple, zagoruyko2015learning, han2015matchnet} (While \cite{santoro2017simple} focuses on reasoning about relation between two objects in a same image which is to address a different problem.). Compared to \cite{koch2015siamese,snell2017prototypical}, this can be seen as providing a learnable rather than fixed metric, or non-linear rather than linear classifier. Compared to \cite{koch2015siamese} we benefit from an episodic training strategy  with an end-to-end manner from scratch, and compared to \cite{santoro2016meta} we avoid the complexity of set-to-set RNN embedding of the sample-set, and simply rely on pooling \cite{santoro2017simple}.

\keypoint{Zero-Shot Learning} Our approach is designed for few-shot learning, but elegantly spans the space into zero-shot learning (ZSL) by modifying the sample branch to input a single category description rather than single training image. When applied to ZSL our architecture is related to methods that learn to align images and category embeddings and perform recognition by predicting if an image and category embedding pair match \cite{frome2013devise,akata2015evaluation,yang2014unified,zhang2015zero}. Similarly to the case with the prior metric-based few-shot approaches, most of these apply a fixed manually defined similarity metric or linear classifier after combining the image and category embedding. In contrast, we again benefit from a deeper end-to-end architecture including a learned non-linear metric in the form of our learned convolutional relation network; as well as from an episode-based training strategy.

\section{Methodology}

\begin{figure*}
\begin{center}

\includegraphics[width=0.75\textwidth]{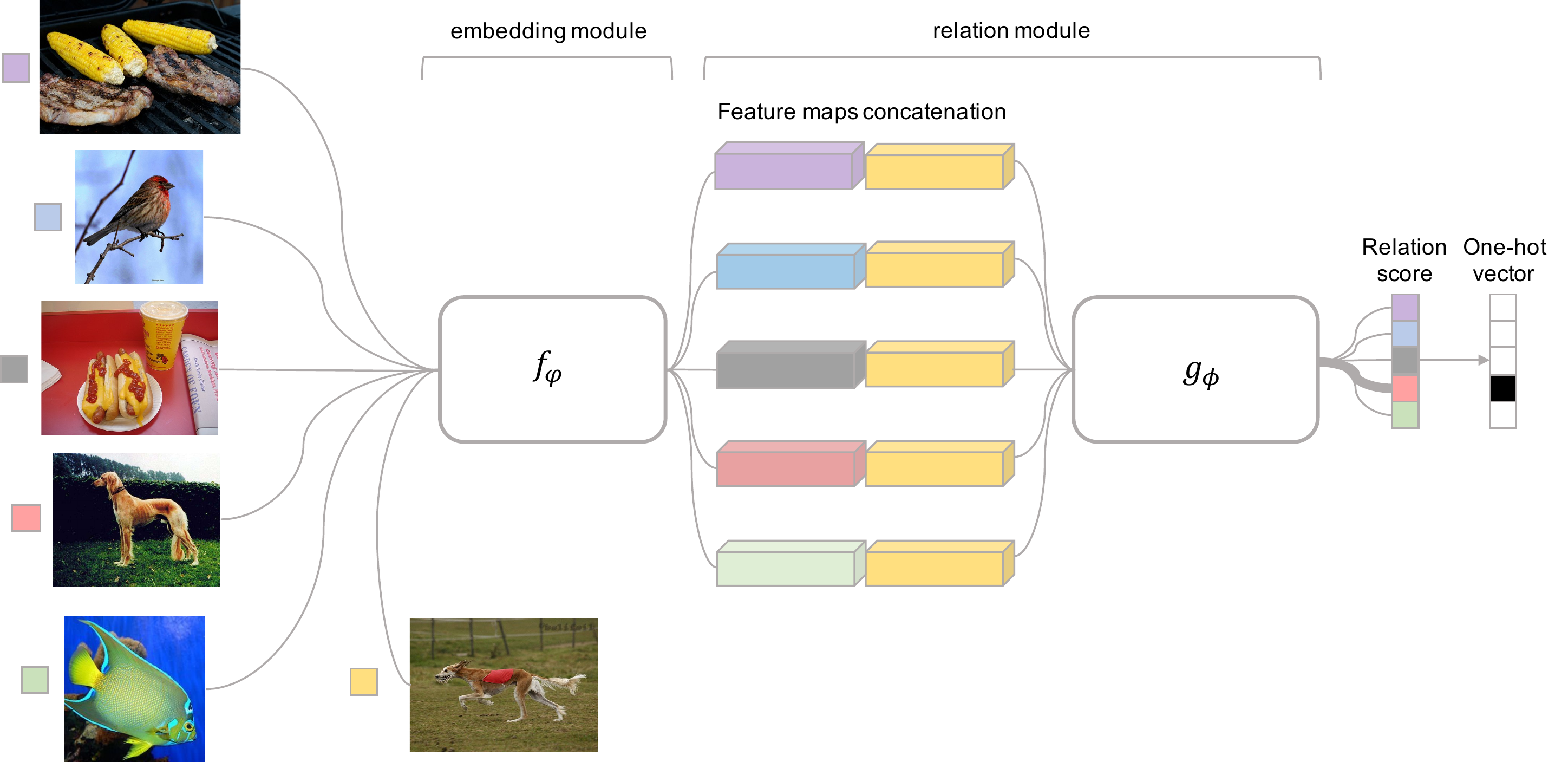}
\end{center}
\caption{\small Relation Network architecture for a 5-way 1-shot problem with one query example.}
\label{fig:schematic}
\vspace{-1em}
\end{figure*}

\subsection{Problem Definition}
We consider the task of few-shot classifier learning. Formally, we have three datasets: a training set, a support set, and a testing set. The  support set and testing set share the same label space, but the training set has its own label space that is disjoint with support/testing set. If the support set contains $K$ labelled examples for each of $C$ unique classes, the target few-shot problem is called $C$-way $K$-shot. 

With the support set only, we can in principle train a classifier to assign a class label $\hat{y}$ to each sample $\hat{x}$ in the test set. However, due to the lack of labelled samples in the support set, the performance of such a classifier is usually not satisfactory. Therefore we aim to perform meta-learning on the training set, in order to extract transferrable knowledge that will allow us to perform better few-shot learning on the support set and thus classify the test set more successfully.

An effective way to exploit the training set is to mimic the few-shot learning setting via {\em episode} based training, as proposed in \cite{vinyals2016matching}. In each training iteration, an episode is formed by randomly selecting $C$ classes from the training set with $K$ labelled samples from each of the $C$ classes to act as the \emph{sample} set $\mathcal{S} = \{ (x_{i}, y_{i})\}^{m}_{i=1}$ ($m = K \times C$), as well as a fraction 
of the remainder of those $C$ classes' samples to serve as the \emph{query} set $\mathcal{Q} = \{ (x_{j}, y_{j})\}^{n}_{j=1}$. This sample/query set split is designed to simulate the support/test set that will be encountered at test time. A model trained from sample/query set can be further fine-tuned using the support set, if desired. In this work we adopt such an episode-based training strategy. In our few-shot experiments (see Section \ref{sec:exp-few-shot}) we consider one-shot ($K=1$, Figure~\ref{fig:schematic}) and five-shot ($K=5$) settings. We also address the $K=0$ zero-shot learning case as explained in Section~\ref{method:ZSL}.


\subsection{Model}
\keypoint{One-Shot}
Our Relation Network (RN) consists of two modules: an {\em embedding} module $f_{\varphi}$ and a {\em relation} module $g_{\phi}$, as illustrated in Figure~\ref{fig:schematic}. Samples $x_{j}$ in the query set $\mathcal{Q}$, and samples $x_i$ in the sample set $\mathcal{S}$ are fed through the embedding module $f_{\varphi}$, which produces feature maps  $f_{\varphi}(x_{i})$ and $f_{\varphi}(x_{j})$. The feature maps $f_{\varphi}(x_{i})$ and $f_{\varphi}(x_{j})$ are combined with operator $\mathcal{C}(f_{\varphi}(x_{i}),f_{\varphi}(x_{j}))$. In this work we assume $\mathcal{C}(\cdot,\cdot)$ to be concatenation of feature maps in depth, although other choices are possible.

The combined feature map of the sample and query are fed into the relation module $g_{\phi}$, which eventually produces a scalar in range of $0$ to $1$ representing the similarity between $x_i$ and $x_j$, which is called relation score. Thus, in the $C$-way one-shot setting, we generate $C$ relation scores $r_{i,j}$ for the relation between one query input $x_j$ and training sample set examples $x_i$,

\begin{equation}
r_{i,j} = g_{\phi} (\mathcal{C}(f_{\varphi}(x_{i}), f_{\varphi}(x_{j}))), \quad i=1,2,\dots,C
\end{equation}

\keypoint{K-shot}
For $K$-shot where $K>1$, we {element-wise sum} over the embedding module outputs of all samples from each training class to form this class' feature map. This pooled class-level feature map is combined with the query image feature map as above. Thus, the number of relation scores for one query is always $C$ in both one-shot or few-shot setting.


\keypoint{Objective function}
We use mean square error (MSE) loss (Eq. (\ref{loss})) to train our model, regressing the relation score $r_{i,j}$ to the ground truth:  matched pairs have similarity $1$ and the mismatched pair have similarity $0$. 

\begin{equation}\label{loss}
\varphi, \phi \leftarrow  \underset{\varphi, \phi}{\operatorname{argmin}}~ \sum_{i=1} ^{m}\sum_{j=1} ^{n}(r_{i,j} - \mathbf{1}(y_{i}==y_{j}))^{2}
\end{equation}

{The choice of MSE  is somewhat non-standard. Our problem may seem to be a classification problem with a label space $\{0,1\}$. However conceptually we are predicting relation scores, which can be considered a regression problem despite that for ground-truth we can only automatically generate $\{0,1\}$ targets. }

\subsection{Zero-shot Learning}\label{method:ZSL}
Zero-shot learning is analogous to one-shot learning in that one datum is given to define each class to recognise. However instead of being given a support set with one-shot image for each of $C$ training classes, it contains a semantic class embedding vector $v_{c}$ for each.
Modifying our framework to deal with the zero-shot case is straightforward: as a different modality of semantic vectors is used for the {\em support} set (\eg attribute vectors instead of images), we use a second heterogeneous embedding module  $f_{\varphi_{2}}$ besides the embedding module $f_{\varphi_{1}}$ used for the image {\em query} set. Then the relation net $g_\phi$ is applied as before. Therefore, the relation score for each query input $x_{j}$ will be:
\begin{equation}
r_{i,j} = g_{\phi} (\mathcal{C}(f_{\varphi_{1}}(v_{c}), f_{\varphi_{2}}(x_{j}))), \quad i=1,2,\dots,C
\end{equation}

The objective function for zero-shot learning is the same as that for few-shot learning.

\subsection{Network Architecture}\label{method:architecture}

\begin{figure}[t]
\begin{center}
\includegraphics[width=0.32\textwidth]{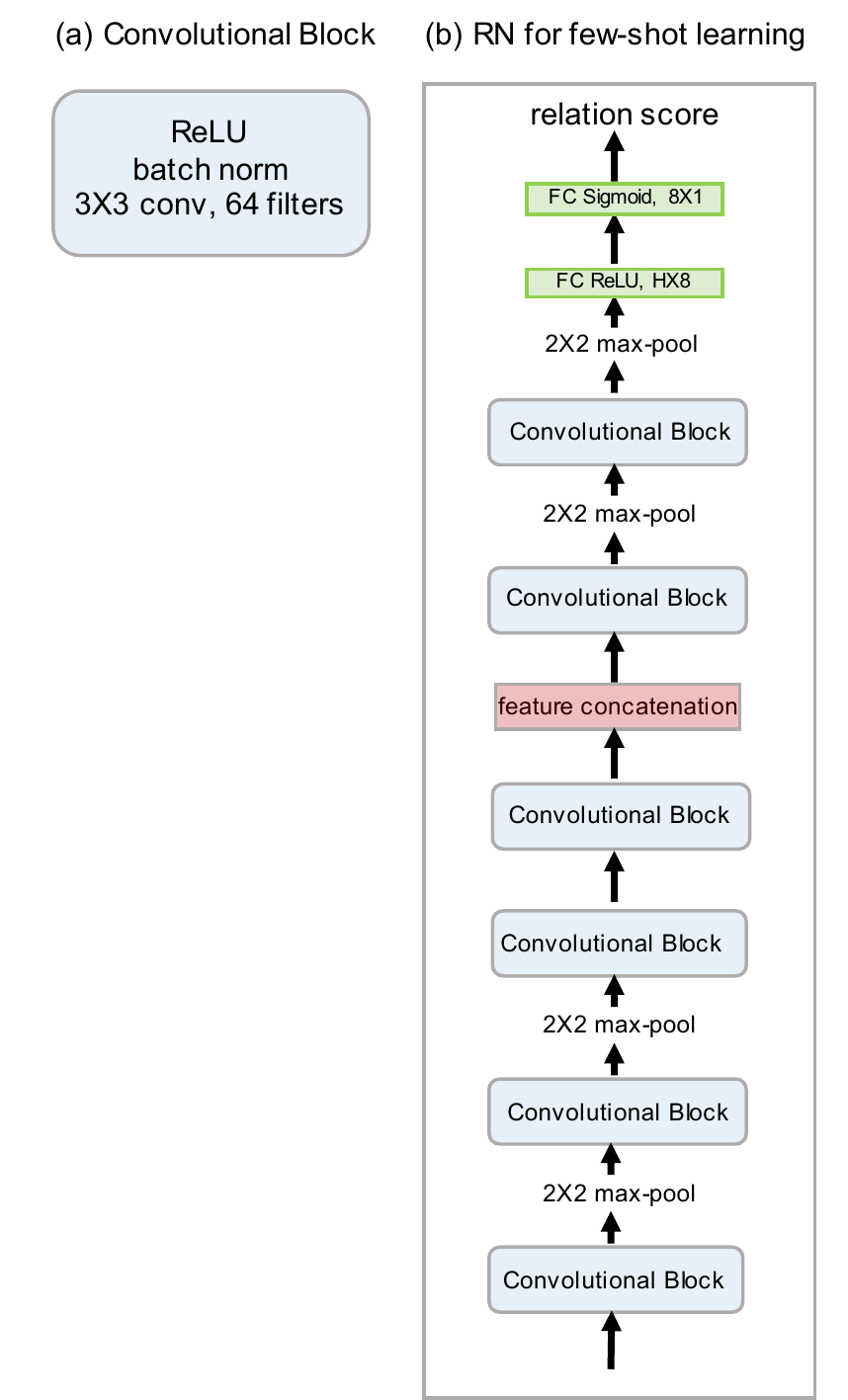}
\end{center}
   \caption{\small Relation Network architecture for few-shot learning (b) which is composed of elements including convolutional block (a).}
\label{fig:networks}
\vspace{-1em}
\end{figure}

As most few-shot learning models utilise four convolutional blocks for embedding module \cite{vinyals2016matching, snell2017prototypical}, we follow the same architecture setting for fair comparison, see Figure \ref{fig:networks}.
More concretely, each convolutional block contains a 64-filter $3 \times 3$ convolution, a batch normalisation and a ReLU nonlinearity layer respectively.  The first two blocks also contain a $2 \times 2$ max-pooling layer while the latter two do not. We do so because we need the output feature maps for further convolutional layers in the relation module. 
The relation module consists of two convolutional blocks and two fully-connected layers. Each of convolutional block is a  $3 \times 3$ convolution with 64 filters followed by batch normalisation, ReLU non-linearity and $2 \times 2$ max-pooling. The output size of last max pooling layer is $\mathcal{H} = 64$ and $\mathcal{H} = 64*3*3 = 576$ for Omniglot and \textit{mini}ImageNet respectively. The two fully-connected layers are 8 and 1 dimensional, respectively.  
All fully-connected layers are ReLU except the output layer is Sigmoid in order to generate relation scores in a {reasonable} range for all versions of our network architecture. 

The zero-shot learning architecture is shown in Figure~\ref{fig:networks2}. In this architecture, the DNN subnet is an existing network (e.g., Inception or ResNet) pretrained on ImageNet.

\section{Experiments}
We evaluate our approach on two related tasks: few-shot classification on Omniglot and \textit{mini}Imagenet, and zero-shot classification on Animals with Attributes (AwA) and Caltech-UCSD Birds-200-2011 (CUB). All the experiments are implemented based on PyTorch~\cite{PyTorch}.

\subsection{Few-shot Recognition}
\label{sec:exp-few-shot}

\keypoint{Settings}
Few-shot learning in all experiments  uses  Adam~\cite{kingma2014adam} with initial learning rate $10^{-3}$ , annealed by half for every 100,000 episodes. 
All our models are end-to-end trained from scratch with no additional dataset.

\keypoint{Baselines} 
We compare against various state of the art baselines for few-shot recognition, including  neural statistician \cite{edwards2016towards},
Matching Nets with and without fine-tuning \cite{vinyals2016matching},
MANN \cite{santoro2016meta},
Siamese Nets with Memory \cite{kaiser2017learning},
Convolutional Siamese Nets \cite{koch2015siamese},
MAML \cite{finn2017model},
Meta Nets \cite{munkhdalai2017meta},
Prototypical Nets \cite{snell2017prototypical} and
Meta-Learner LSTM \cite{ravi2016optimization}.

\subsubsection{Omniglot}
\keypoint{Dataset}
 Omniglot \cite{lake2011one} contains 1623 characters (classes) from 50 different alphabets. Each class contains 20 samples drawn by different people.
Following \cite{santoro2016meta, vinyals2016matching, snell2017prototypical}, we augment new classes through $90^{\circ}$, $180^{\circ}$ and $270^{\circ}$ rotations of existing data and use 1200 original classes plus rotations for training and remaining 423 classes plus rotations for testing. All input images are resized to $28 \times 28$. 


\keypoint{Training} 
Besides the $K$ sample images, the \textbf{5-way 1-shot} contains 19 query images, the \textbf{5-way 5-shot} has 15 query images, the \textbf{20-way 1-shot} has 10 query images and the \textbf{20-way 5-shot} has 5 query images for each of the $C$ sampled classes in each training episode. This means for example that there are  $19 \times 5 + 1 \times 5 = 100$ images in one training episode/mini-batch for the 5-way 1-shot experiments.

\keypoint{Results} 
Following \cite{snell2017prototypical}, we computed few-shot classification accuracies on Omniglot by averaging over 1000 randomly generated episodes from the testing set. 
For the 1-shot and 5-shot experiments, we batch one and five query images per class respectively for evaluation during testing.
The results are shown in Table \ref{tab:omni}. We achieved state-of-the-art performance under all experiments setting with higher averaged accuracies and lower standard deviations, except 5-way 5-shot where our model is 0.1\% lower in accuracy than \cite{finn2017model}. This is despite that many alternatives have significantly more complicated machinery \cite{munkhdalai2017meta,edwards2016towards}, or fine-tune on the target problem \cite{finn2017model,vinyals2016matching},  while we do not.

\setlength{\tabcolsep}{10pt}
\begin{table*}[t]
\centering
\footnotesize
\begin{tabular}{@{} lccccc @{}}
\toprule
\multirow{2}{*}{\bf Model} & \multirow{2}{*}{\bf Fine Tune} &\multicolumn{2}{c}{\multirow{2}{*}{\bf 5-way Acc.}} &\multicolumn{2}{c}{\multirow{2}{*}{\bf 20-way Acc.}}\\
& \multicolumn{2}{c}{} & \multicolumn{2}{c}{} \\
& & 1-shot & 5-shot & 1-shot &5-shot  \\
\midrule 

\textbf{\textsc{Mann}} \cite{santoro2016meta} & N  & 82.8\% & 94.9\% & - &- \\
\textbf{\textsc{Convolutional} \textsc{Siamese} \textsc{Nets}} \cite{koch2015siamese}& N  &96.7\% &98.4\% &88.0\% &96.5\% \\ 
\textbf{\textsc{Convolutional} \textsc{Siamese} \textsc{Nets}} \cite{koch2015siamese}& Y  &97.3\% &98.4\% &88.1\% &97.0\% \\ 
\textbf{\textsc{Matching} \textsc{Nets}} \cite{vinyals2016matching}& N & 98.1\% & 98.9\% &93.8\% & 98.5\% \\
\textbf{\textsc{Matching} \textsc{Nets}} \cite{vinyals2016matching}& Y &97.9\% &98.7\% &93.5\% &98.7\% \\ 
\textbf{\textsc{Siamese} \textsc{Nets} \textsc{with} \textsc{Memory}} \cite{kaiser2017learning}&N&98.4\% &99.6\% &95.0\% &98.6\% \\
\textbf{\textsc{Neural} \textsc{Statistician}} \cite{edwards2016towards}&  N& 98.1\%& 99.5\%& 93.2\% &  98.1\%\\ 
\textbf{\textsc{Meta} \textsc{Nets}} \cite{munkhdalai2017meta}& N &99.0\% & - & 97.0\% & -\\ 
\textbf{\textsc{Prototypical} \textsc{Nets}} \cite{snell2017prototypical}&N  &98.8\% &99.7\%  &96.0\% &98.9\% \\ 
\textbf{\textsc{Maml}} \cite{finn2017model}& Y &98.7 $\pm$ 0.4\% &\textbf{99.9 $\pm$ 0.1}\% &95.8 $\pm$ 0.3\% &98.9 $\pm$ 0.2\% \\ 
\midrule
\textbf{\textsc{Relation} \textsc{Net}}&N& \textbf{99.6 $\pm$ 0.2\%} &\textbf{99.8$\pm$ 0.1\%} &\textbf{97.6 $\pm$ 0.2\%} &\textbf{99.1$\pm$ 0.1\%}\\ 
\bottomrule
\end{tabular}
\caption{\small 
Omniglot few-shot classification. 
Results are accuracies averaged over 1000 test episodes and with 95\% confidence intervals where reported. 
The best-performing method is highlighted, along with others whose confidence intervals overlap. `-': not reported.
}
\label{tab:omni}
\vspace{-1em}
\end{table*}


\begin{figure}[t]
\begin{center}
\includegraphics[width=0.25\textwidth]{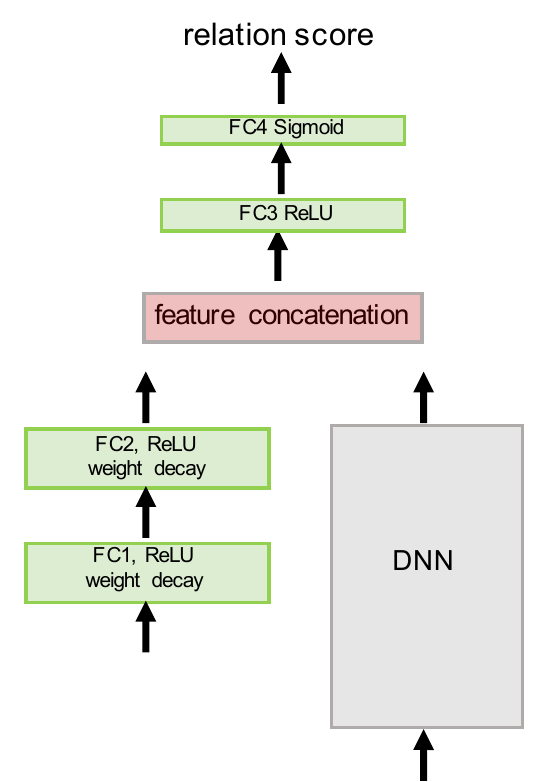}
\end{center}
   \caption{\small Relation Network architecture for zero-shot learning.}
\label{fig:networks2}
\vspace{-1em}
\end{figure}

\subsubsection{\textit{mini}ImageNet} 
\paragraph{Dataset}
The \textit{mini}Imagenet dataset, originally proposed by \cite{vinyals2016matching}, consists of 60,000 colour images with 100 classes, each having 600 examples. 
We followed the split introduced by \cite{ravi2016optimization}, with 64, 16, and 20 classes for training, validation and testing, respectively. The 16 validation classes is used for monitoring generalisation performance only.

\keypoint{Training}
Following the standard setting adopted by most existing few-shot learning work, we conducted 5 way 1-shot and 5-shot classification. 
Beside the $K$ sample images, the \textbf{5-way 1-shot} contains 15 query images, and the \textbf{5-way 5-shot} has 10 query images for each of the $C$ sampled classes in each training episode.
This means for example that there are $15 \times 5 + 1 \times 5 = 80$ images in one training episode/mini-batch for 5-way 1-shot experiments. We resize  input images to $84 \times 84$. Our model is trained end-to-end from scratch, with random initialisation, and no additional training set. 


\keypoint{Results}
Following \cite{snell2017prototypical}, we batch 15 query images per class in each episode for evaluation in both 1-shot and 5-shot scenarios and the few-shot classification accuracies are computed by averaging over 600 randomly generated episodes from the test set.

From Table \ref{tab:mini}, we can see that our model achieved state-of-the-art performance on 5-way 1-shot settings and competitive results on 5-way 5-shot.
However, the 1-shot result reported by prototypical networks \cite{snell2017prototypical} reqired to be trained on 30-way 15 queries per training episode, and 5-shot result was trained on 20-way 15 queries per training episode.
When trained with 5-way 15 query per training episode, \cite{snell2017prototypical} only got $46.14 \pm 0.77\%$ for 1-shot evaluation,
clearly weaker than ours.
In contrast, all our models are trained on 5-way, 1 query for 1-shot and 5 queries for 5-shot per training episode, with much less training queries than \cite{snell2017prototypical}. 


\setlength{\tabcolsep}{4.8pt}
\begin{table}[t]
\centering
\footnotesize
\begin{tabular}{@{} lccc @{}}
\toprule
\multirow{2}{*}{\bf Model} & \multirow{2}{*}{\bf FT} &\multicolumn{2}{c}{\multirow{2}{*}{\bf 5-way Acc.}}\\
& \multicolumn{2}{c}{}  \\
& & 1-shot & 5-shot \\
\midrule 

\textbf{\textsc{Matching} \textsc{Nets}} \cite{vinyals2016matching}& N &43.56 $\pm$ 0.84\% &55.31 $\pm$ 0.73\%  \\ 
\textbf{\textsc{Meta} \textsc{Nets}} \cite{munkhdalai2017meta}& N &49.21 $\pm$ 0.96\% & - \\ 
\textbf{\textsc{Meta}-\textsc{Learn} \textsc{LSTM}} \cite{ravi2016optimization}& N &43.44 $\pm$ 0.77\% & 60.60 $\pm$ 0.71\% \\ 
\textbf{\textsc{Maml}} \cite{finn2017model}& Y& 48.70 $\pm$ 1.84\% & 63.11 $\pm$ 0.92\% \\ 
\textbf{\textsc{Prototypical} \textsc{Nets}} \cite{snell2017prototypical}&N  &49.42 $ \pm $ 0.78\% &\textbf{68.20 $\pm$ 0.66\%}  \\ 
\midrule
\textbf{\textsc{Relation} \textsc{Net}}&N& \textbf{50.44 $\pm$ 0.82\%} &65.32 $\pm$ 0.70\% \\ 
\bottomrule
\end{tabular}%
\caption{\small
Few-shot classification accuracies on \textit{mini}Imagenet. All accuracy results are averaged over 600 test episodes and are reported with 95\% confidence intervals, same as \cite{snell2017prototypical}. For each task, the best-performing method is highlighted, along with any others whose confidence intervals overlap. `-': not reported.
}
\label{tab:mini}
\end{table}

\subsection{Zero-shot Recognition}
\keypoint{Datasets and settings}
We follow two ZSL settings: the \emph{old} setting and the new \emph{GBU} setting provided by~\cite{xian2017zero} for training/test splits. 
Under the \emph{old} setting, adopted by most existing ZSL works before \cite{xian2017zero}, 
some of the test classes also appear in
the ImageNet 1K classes, which have been used to pretrain the image embedding 
network, thus violating the zero-shot assumption.
In contrast, the new \emph{GBU} setting ensures that none of the test classes of the datasets appear in
the ImageNet 1K classes. Under both settings, the test set can comprise only the unseen class samples (conventional test set setting) or a mixture of seen and unseen class samples. The latter, termed   generalised zero-shot learning (GZSL), is more realistic in practice.  

Two widely used ZSL benchmarks are selected for the \emph{old} setting:
\textbf{AwA} (Animals with Attributes)~\cite{lampert2014attribute} consists of 30,745 images of 50 classes of animals. It has a fixed split for evaluation with 40 training classes and 10 test classes. \textbf{CUB} (Caltech-UCSD Birds-200-2011)~\cite{wah2011multiclass} contains 11,788 images of 200 bird species with 150 seen classes and 50 disjoint unseen classes. 
Three datasets~\cite{xian2017zero} are selected for \emph{GBU} setting: \textbf{AwA1}, \textbf{AwA2} and \textbf{CUB}. The newly released AwA2~\cite{xian2017zero} consists of 37,322 images of 50 classes which is an extension of AwA while AwA1 is same as AwA but under the \emph{GBU} setting.

\keypoint{Semantic representation}
For \textbf{AwA}, we use the continuous 85-dimension class-level attribute vector from \cite{lampert2014attribute}, which has been used by all recent works. For~\textbf{CUB}, a continuous 312-dimension class-level attribute vector is used.

\keypoint{Implementation details}
Two different embedding modules are used for the two input modalities in zero-shot learning. 
Unless otherwise specified, we use Inception-V2~\cite{szegedy2015going, ioffe2015batch} as the query image embedding DNN   in the old and conventional setting and ResNet101 \cite{he2016cvpr} for the GBU and generalised setting, taking the top pooling units as image embedding with dimension $D=1024$ and $2048$ respectively. This DNN is pre-trained on ILSVRC 2012 1K classification without fine-tuning,  as in recent deep ZSL works~\cite{lei2015predicting, reed2016learning, zhang2017learning}. A MLP network is used for embedding semantic attribute vectors. The size of  hidden layer FC1 (Figure \ref{fig:networks2}) is set to 1024 and 1200 for AwA and CUB respectively, and the output size FC2 is set to the same dimension as the image embedding for both datasets. For the relation module, the image and semantic embeddings are concatenated before being fed into MLPs with hidden layer FC3 size 400 and 1200 for AwA and CUB, respectively. 

We add weight decay (L2 regularisation) in FC1 \& 2 as there is a hubness problem~\cite{zhang2017learning} in cross-modal mapping
for ZSL which can be best solved by mapping the semantic feature vector to the visual feature space with regularisation. After that, FC3 \& 4 (relation module) are used to compute the relation between the semantic representation (in the visual feature space) and the visual representation. Since the hubness problem does not existing in this step, no L2 regularisation/weight decay is needed.
All the ZSL models are trained with weight decay $10^{-5}$ in the embedding network. The learning rate is initialised to $10^{-5}$ with Adam~\cite{kingma2014adam} and then annealed by half every 200,000 iterations.

\keypoint{Results under the old setting}
The conventional evaluation for ZSL followed by the majority of prior work is to assume that the test data all comes from unseen classes. We evaluate this setting first. 
We compare 15 alternative approaches in Table~\ref{tab:zsl}. With only the attribute vector used as the sample class embedding, our model achieves competitive result on AwA and state-of-the-art performance on the more challenging CUB dataset, outperforming the most related alternative prototypical networks \cite{snell2017prototypical} by a big margin.
Note that only inductive methods are considered. Some recent methods \cite{zhang2016zeroshot, fu2014transductive, fu2016semi} are tranductive in that they use all test data at once for model training, which gives them a big advantage at the cost of making a very strong assumption that may not be met in practical applications, so we do not compare with them here.

\setlength{\tabcolsep}{6pt}
\begin{table}[t]
\centering
\footnotesize
\begin{tabular}{@{} lcccc @{}}
\toprule

\textbf{Model} & \textbf{F}     & \textbf{SS}   & \textbf{AwA} & \textbf{CUB} \\ 
 &     &   & 10-way 0-shot & 50-way 0-shot \\ 
\midrule 

\textbf{\textsc{Sje}}~\cite{akata2015evaluation}&$F_{G}$ & A & 66.7 &50.1  \\
\textbf{\textsc{Eszsl}}~\cite{romera2015embarrassingly}&$F_{G}$ & A & 76.3 & 47.2 \\
\textbf{\textsc{Sse}-\textsc{Relu}}~\cite{zhang2015zero}&$F_{V}$ & A & 76.3 & 30.4 \\
\textbf{\textsc{Jlse}}~\cite{zhang2016zero}&$F_{V}$  & A & 80.5 & 42.1 \\ 
\textbf{\textsc{Sync-struct}}~\cite{changpinyo2016synthesized}&$F_{G}$ & A & 72.9 & 54.5 \\ 
\textbf{\textsc{Sec-ml}}~\cite{bucher2016improving}&$F_{V}$  & A & 77.3 & 43.3 \\  
\textbf{\textsc{Proto.} \textsc{Nets}}~\cite{snell2017prototypical}&$F_{G}$ & A & - & 54.6\\ 
\midrule
\textbf{\textsc{Devise}}~\cite{frome2013devise}&$N_{G}$ & A/W &56.7/50.4  &33.5   \\ 
\textbf{\textsc{Socher} {\em et al.}}~\cite{socher2013zero}&$N_{G}$ & A/W & 60.8/50.3 &39.6  \\ 
\textbf{\textsc{Mtmdl}}~\cite{yang2014unified}&$N_{G}$ & A/W &63.7/55.3   &32.3   \\ 
\textbf{\textsc{Ba} {\em et al.}}~\cite{lei2015predicting}&$N_{G}$ & A/W &69.3/58.7  &34.0   \\ 
\textbf{\textsc{Ds-sje}}~\cite{reed2016learning}&$N_{G}$ & A/D & - & 50.4/ 56.8  \\ 
\textbf{\textsc{Sae}}~\cite{kodirov2017semantic}&$N_{G}$ & A & 84.7 & 61.4 \\
\textbf{\textsc{Dem}}~\cite{zhang2017learning}&$N_{G}$ & A/W & \textbf{86.7}/78.8 & 58.3 \\ 
\midrule
\textbf{\textsc{Relation} \textsc{Net}}&$N_{G}$ & A & 84.5 &\textbf{62.0}\\ 
\bottomrule
\end{tabular}
\caption{\footnotesize Zero-shot classification accuracy (\%) comparison on AwA and CUB (hit@1 accuracy over all samples) under the old and conventional setting. SS: semantic space; A: attribute space; W: semantic word vector space; D: sentence description (only available for CUB). F: how the visual feature space is computed; For non-deep models: $F_O$ if overfeat \cite{sermanet2013overfeat} is used; $F_G$ for GoogLeNet \cite{szegedy2015going}; and $F_V$ for VGG net \cite{simonyan2014very}. For neural network based methods, all use Inception-V2 (GoogLeNet with batch normalisation)~\cite{szegedy2015going, ioffe2015batch} as the DNN image imbedding subnet, indicated as $N_G$.}
\label{tab:zsl}
\end{table}

\keypoint{Results under the GBU setting} 
We follow the  evaluation setting of \cite{xian2017zero}. 
We compare our model with 11 alternative ZSL models in Table \ref{tab:gzsl}. The 10 shallow models’ results are from \cite{xian2017zero} and the result of the state-of-the-art method DEM ~\cite{zhang2017learning} is from the authors' GitHub page\footnote{\url{https://github.com/lzrobots/DeepEmbeddingModel_ZSL}}.
We can see that on AwA2 and CUB, Our model is particularly
strong under the more realistic GZSL setting measured using the harmonic mean
(H) metric. While on AwA1, our method is only outperformed by DEM~\cite{zhang2017learning}.

\setlength{\tabcolsep}{9pt}
\begin{table*}[ht]
\centering
\footnotesize
\begin{tabular}{@{} l|c|ccc|c|ccc|c|ccc @{}}
\toprule
&\multicolumn{4}{c|}{\bf AwA1}&\multicolumn{4}{c|}{\bf AwA2}&\multicolumn{4}{c}{\bf CUB}\\\midrule
&\bf ZSL & \multicolumn{3}{c|}{\bf GZSL} &\bf ZSL & \multicolumn{3}{c|}{\bf GZSL} &\bf ZSL & \multicolumn{3}{c}{\bf GZSL}\\
Model &\textbf{T1} & \textbf{u} & \textbf{s} & \textbf{H}&\textbf{T1} & \textbf{u} & \textbf{s} & \textbf{H} & \textbf{T1} & \textbf{u} & \textbf{s} & \textbf{H} \\
\midrule 
\textbf{\textsc{Dap}}~\cite{lampert2014attribute} &44.1 &0.0&88.7  &0.0 &46.1 &0.0 &84.7 &0.0 &40.0 &1.7& 67.9 &3.3\\ 
\textbf{\textsc{Conse}}~\cite{norouzi2013zero} &45.6 &0.4&88.6  &0.8 &44.5 &0.5 & 90.6&1.0 &34.3 &1.6& \textbf{72.2} &3.1\\ 
\textbf{\textsc{Sse}}~\cite{zhang2015zero} &60.1 &7.0& 80.5 &12.9 &61.0 &8.1 &82.5 &14.8 &43.9 &8.5&46.9  &14.4\\ 
\textbf{\textsc{Devise}}~\cite{frome2013devise} &54.2 &  13.4& 68.7 &22.4 &59.7 &17.1 & 74.7& 27.8&52.0 &  23.8& 53.0 &32.8\\ 
\textbf{\textsc{Sje}}~\cite{akata2015evaluation}  &65.6 &  11.3 &74.6 &19.6 &61.9 &8.0 &73.9 & 14.4&53.9 &  23.5 &59.2&33.6\\ 
\textbf{\textsc{Latem}}~\cite{xian2016latent}  &55.1 & 7.3 &71.7 &13.3 &55.8 &11.5 &77.3 & 20.0&49.3 & 15.2 &57.3 &24.0\\ 
\textbf{\textsc{Eszsl}}~\cite{romera2015embarrassingly}  &58.2 &  6.6& 75.6& 12.1 &58.6 &5.9 & 77.8& 11.0&53.9 &  12.6 &63.8&21.0\\ 
\textbf{\textsc{Ale}}~\cite{akata2016label}  &59.9 & 16.8& 76.1& 27.5 &62.5 &14.0 &81.8 &23.9 &54.9 & 23.7& 62.8& 34.4 \\ 
\textbf{\textsc{Sync}}~\cite{changpinyo2016synthesized}  &54.0 & 8.9& 87.3& 16.2 &46.6 &10.0 & 90.5&18.0 &55.6 & 11.5& 70.9& 19.8\\ 
\textbf{\textsc{Sae}}~\cite{kodirov2017semantic}  &53.0 & 1.8& 77.1& 3.5 &54.1 &1.1 & 82.2&2.2 &33.3 & 7.8& 57.9& 29.2\\ 
\midrule
\textbf{\textsc{Dem}}~\cite{zhang2017learning}  &\textbf{68.4} & \textbf{32.8}& 84.7&\textbf{47.3} &\textbf{67.1} &\textbf{30.5} & 86.4&45.1 &51.7 & 19.6& 54.0& 13.6\\ 
\midrule
\textbf{\textsc{Relation} \textsc{Net}} &68.2 &31.4 &\textbf{91.3} &46.7 &64.2 &30.0 &\textbf{93.4} &\textbf{45.3}  &\textbf{55.6} &\textbf{38.1} &61.1 &\textbf{47.0}  \\ 
\bottomrule
\end{tabular}%
\vspace{-0.2cm}
\caption{\small Comparative results under the GBU setting. Under the conventional 
ZSL setting, the performance is evaluated using  per-class average Top-1 (\textbf{T1}) accuracy (\%), and under GZSL, it is  measured using  \textbf{u} = \textbf{T1} on unseen classes, \textbf{s} = \textbf{T1} on seen classes, and \textbf{H} = harmonic mean.}
\label{tab:gzsl}
\end{table*}

\section{Why does Relation Network Work?}\label{model_sanity}
\subsection{Relationship to existing models}
Related prior few-shot work uses fixed pre-specified distance metrics such as Euclidean or cosine distance to perform classification \cite{vinyals2016matching, snell2017prototypical}. These studies can be seen as distance metric learning, but where all the learning occurs in the feature embedding, and a fixed metric is used given the learned embedding. Also related are conventional metric learning approaches \cite{mensink2012metric,chen2012jointBayesian} that focus on learning a shallow (linear) Mahalanobis metric for a fixed feature representation. In contrast to prior work's fixed metric or fixed features and shallow learned metric, Relation Network can be seen as  both learning a deep embedding {\em and} learning a deep non-linear metric (similarity function)\footnote{Our architecture does not guarantee the self-similarity and symmetry properties of a formal similarity function. But empirically we find these properties hold numerically for a trained Relation Network.}. These are mutually tuned end-to-end to support each other in few short learning.

Why might this be particularly useful? By using a flexible function approximator to learn similarity, we learn a good metric in a data driven way and do not have to manually choose the right metric (Euclidean, cosine, Mahalanobis). Fixed metrics like \cite{vinyals2016matching, snell2017prototypical} assume that features are solely compared element-wise, and the most related \cite{snell2017prototypical} assumes linear separability after the embedding. 
These are thus critically dependent on the efficacy of the learned embedding network, and hence limited by the extent to which the embedding networks generate inadequately discriminative representations. 
In contrast, by deep learning a  non-linear similarity metric jointly with the embedding, Relation Network can better identify matching/mismatching pairs. 

\begin{figure}[t] 
	\begin{subfigure}[b]{0.5\linewidth}
		\centering
		\includegraphics[width=0.99\linewidth]{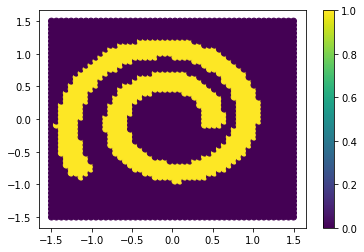} 
		\caption{Ground Truth} 
		\label{fig8:a} 
	\end{subfigure}
	\begin{subfigure}[b]{0.5\linewidth}
		\centering
		\includegraphics[width=0.99\linewidth]{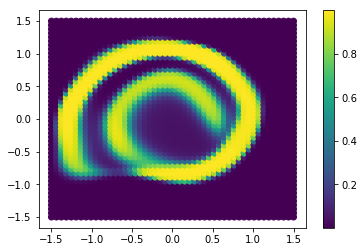} 
		\caption{Relation Network} 
		\label{fig8:b} 
	\end{subfigure} 
	\begin{subfigure}[b]{0.5\linewidth}
		\centering
		\includegraphics[width=0.99\linewidth]{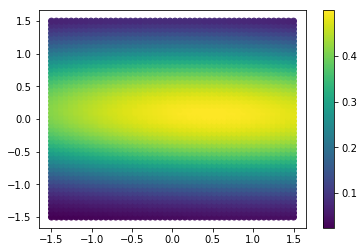} 
		\caption{Metric Learning} 
		\label{fig8:c} 
	\end{subfigure}
	\begin{subfigure}[b]{0.5\linewidth}
		\centering
		\includegraphics[width=0.99\linewidth]{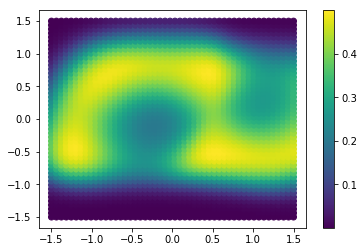} 
		\caption{Metric + Embedding} 
		\label{fig8:d} 
	\end{subfigure} 
	\caption{\small An example relation learnable by Relation Network and not by non-linear embedding + metric learning.}
	\label{fig8} 
\end{figure}

\subsection{Visualisation}
To illustrate the previous point about adequacy of learned input embeddings, we show a synthetic example where existing approaches definitely fail and our Relation Network can succeed due to using a deep relation module. Assuming 2D query and sample input embeddings to a relation module, Fig.~\ref{fig8}(a) shows the space of 2D sample inputs for a fixed 2D query input. Each sample input (pixel) is colored according to whether it matches the fixed query or not. This represents a case where the output of the embedding modules is not discriminative enough for trivial (Euclidean NN) comparison between query and sample set. In Fig.~\ref{fig8}(c) we attempt to learn matching via a Mahalanobis metric learning relation module, and we can see the result is inadequate. In Fig.~\ref{fig8}(d) we  learn a further 2-hidden layer MLP embedding of query and sample inputs as well as the subsequent Mahalanobis metric, which is also not adequate. Only by learning the full deep relation module for similarity can we solve this problem in Fig.~\ref{fig8}(b).

\begin{figure}[t] 
\centering
\includegraphics[width=0.48\columnwidth]{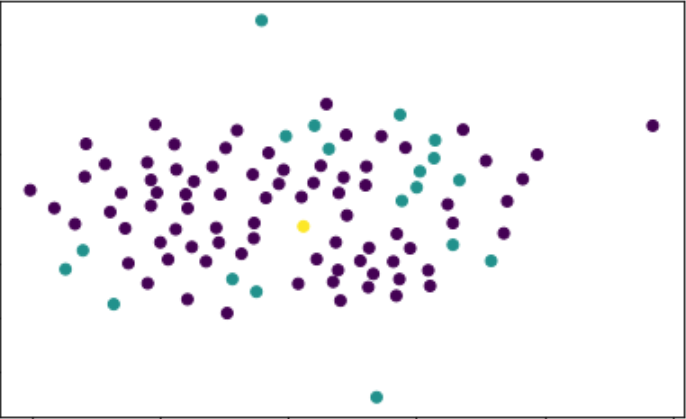} 
~
\includegraphics[width=0.48\columnwidth]{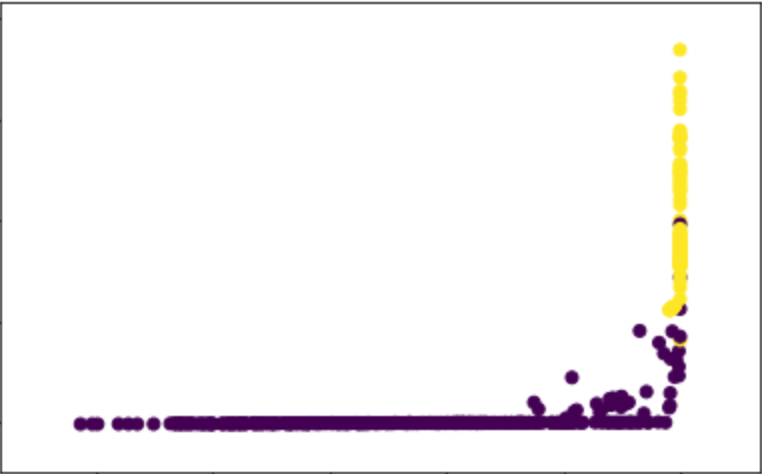} 
\\
\includegraphics[width=0.48\columnwidth]{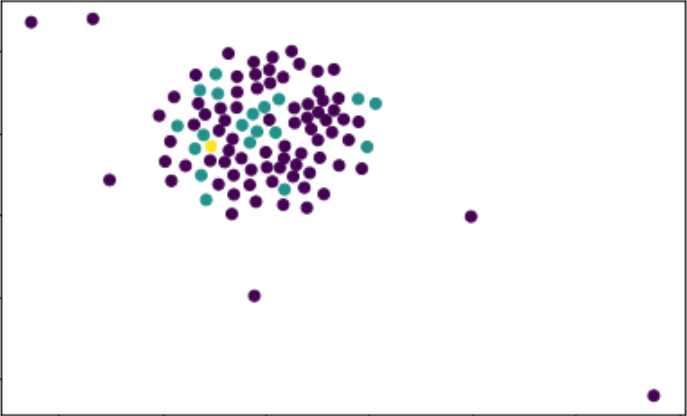} 
~
\includegraphics[width=0.48\columnwidth]{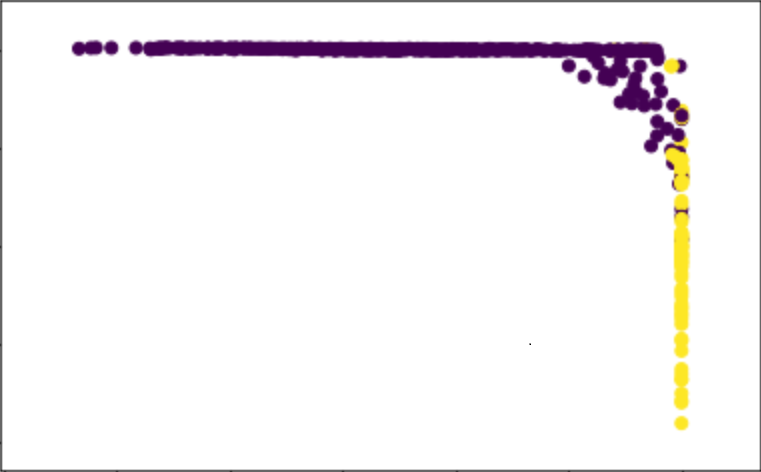} 
\caption{\small Example Omniglot few-shot problem visualisations. Left: Matched (cyan) and mismatched (magenta) sample embeddings for a given query (yellow) are not straightforward to differentiate. Right: Matched (yellow) and mismatched (magenta) relation module pair representations are linearly separable. }
\label{fig:Embed} 
\end{figure}

In a real problem the difficulty of comparing embeddings may not be this extreme, but it can still be challenging. We qualitatively illustrate the challenge of matching two example Omniglot query images (embeddings projected to 2D, Figure~\ref{fig:Embed}(left)) by showing an analogous plot of real sample images colored by match (cyan) or mismatch (magenta) to two example queries (yellow). 
Under standard assumptions \cite{vinyals2016matching, snell2017prototypical,mensink2012metric,chen2012jointBayesian} the cyan matching samples should be nearest neighbours to the yellow query image with some metric (Euclidean, Cosine, Mahalanobis). But we can see that the match relation is more complex than this.  In Figure~\ref{fig:Embed}(right), we instead plot the same two example queries in terms of a 2D PCA representation of each query-sample pair, as represented by the relation module's penultimate layer. We can see that the relation network has mapped the data into a space where the (mis)matched pairs are linearly separable.

%
%
%
%


%

\section{Conclusion}
We proposed a simple method called the Relation Network for few-shot and zero-shot learning. Relation network learns an embedding and a deep non-linear distance metric for comparing query and sample items. Training the network end-to-end with episodic training tunes the embedding and distance metric for effective few-shot learning. This approach is far simpler and more efficient than recent few-shot meta-learning approaches, and produces state-of-the-art results. It further proves effective at both conventional and generalised zero-shot settings. 


\keypoint{Acknowledgements}
This work was supported by the ERC grant ERC-2012-AdG 321162-HELIOS, EPSRC grant Seebibyte EP/M013774/1, EPSRC/MURI grant EP/N019474/1, EPSRC grant EP/R026173/1, and the European Union's Horizon 2020 research and innovation program (grant agreement no. 640891). We gratefully acknowledge the support of NVIDIA Corporation with the donation of the Titan Xp GPU and the ESPRC funded Tier 2 facility, JADE used for this research.
{\small
\bibliographystyle{ieee}
\bibliography{egbib}
}

\end{document}